 \documentclass[sigconf,screen]{acmart}

\usepackage{algorithm}
\usepackage{algorithmic}
\usepackage{subcaption}
\usepackage{comment}
\AtBeginDocument{%
  \providecommand\BibTeX{{%
    \normalfont B\kern-0.5em{\scshape i\kern-0.25em b}\kern-0.8em\TeX}}}

\acmConference[SIGKDD '20]{SIGKDD '20: ACM Conference on Knowledge Discovery and Data Mining}{August 22--27, 2020}{San Diego, CA}

\begin{document}

\title{FedCD: Improving Performance in non-IID Federated Learning}


\author{Kavya Kopparapu}
\affiliation{%
  \institution{Harvard University}
  \city{Cambridge}
  \country{USA}
}
\email{kavyakopparapu@college.harvard.edu}
\authornote{All authors contributed equally to this research and are listed in alphabetical order.}

\author{Eric Lin}
\affiliation{%
  \institution{Harvard University}
  \city{Cambridge}
  \country{USA}
}
\email{eric\_lin@college.harvard.edu}
\authornotemark[1]

\author{Jessica Zhao}
\affiliation{%
  \institution{Harvard University}
  \city{Cambridge}
  \country{USA}
}
\email{jzhao@college.harvard.edu}
\authornotemark[1]

\renewcommand{\shortauthors}{Kopparapu, Lin, and Zhao.}

\begin{abstract}
  A clear and well-documented \LaTeX\ document is presented as an
  article formatted for publication by ACM in a conference proceedings
  or journal publication. Based on the ``acmart'' document class, this
  article presents and explains many of the common variations, as well
  as many of the formatting elements an author may use in the
  preparation of the documentation of their work.
\end{abstract}



\begin{abstract}
Federated learning has been widely applied to enable decentralized devices, which each have their own local data, to learn a shared model. However, learning from real-world data can be challenging, as it is rarely identically and independently distributed (IID) across edge devices (a key assumption for current high-performing and low-bandwidth algorithms). We present a novel approach, FedCD, which clones and deletes models to dynamically group devices with similar data. Experiments on the CIFAR-10 dataset show that FedCD achieves higher accuracy and faster convergence compared to a FedAvg baseline on non-IID data while incurring minimal computation, communication, and storage overheads. 
\end{abstract}

\maketitle

\section{Introduction}
The most successful machine learning methods generalize well to different data sources by training on large amounts of data. However, in many important applications such as healthcare, data is subject to strict privacy constraints that prevent direct access to local data. In addition, devices often have limited communication bandwidth and on-device memory.


Federated learning is an increasingly popular method that addresses these constraints. In particular, it differs from other machine learning approaches by allowing multiple edge devices to learn a shared global model without the need to reveal their data to the central server. 
Under the standard federated learning approach (FedAvg), each device trains a copy of the global model locally on its own data and sends a weight update to the central server, which averages all updated model weights and re-deploys them as the new model to the individual devices \cite{fedavg}. This allows a single global model to train on multiple devices' sensitive data without compromising privacy, e.g. by moving the data off-device.

Unfortunately, FedAvg and other recently developed privacy- and bandwidth-conscious approaches perform poorly when data is not independent and identically distributed (IID) across devices \cite{KairouzPeter2019AaOP}.
Non-IID data may cause different devices' updates to conflict with each other, which could lead to significant oscillations between training rounds and slower convergence. 

Devices often belong to one of many archetypes, where an archetype describes a subset of non-IID data that is itself IID. Previously proposed learning schemes such as FedAvg attempt to learn a single global model that performs well for all archetypes, yet this is often difficult or even infeasible when data is non-IID. 
In contrast, we propose Federated Cloning-and-Deletion (FedCD), a learning scheme that results in a specialized model for each archetype through iterative cloning of global models at specified milestones, adaptive updating of a high-scoring subset of global models, and deletion of poor-performing models. By maintaining multiple global models, devices can preferentially update models that perform well on their local data, thus self-selecting into groups with similar data. This allows for both faster convergence and higher accuracy.


\subsection{Related Work}
Most federated learning approaches use stochastic gradient descent, which optimally requires IID sampling of the data. In practice, federated learning rarely sees IID data across edge devices and learning on non-IID data is an open problem \cite{KairouzPeter2019AaOP}. Recent work has proposed various solutions to addressing this challenge:

\subsubsection{Globally Shared Subsets} Zhao et al. found that sharing just 5\% of global data improved accuracy by 30\% on non-iid subsets of the CIFAR-10 dataset \cite{globaldata}.  However, a globally shared subset of data that is representative of all devices' individual data can be difficult to obtain or synthesize and is generally infeasible in many contexts. 

\subsubsection{Peer-to-peer Federated Learning} Peer-to-peer learning schemes increase the number of global models and the communication cost per round as every device participates in each round with a unique model \cite{edgify, BelletGTT17}. Although this approach increases accuracy, in many scenarios such as deploying edge devices in the field where security is important, individual learners would not be connected to each other in favor of maintaining a single, stable connection to a centralized server. Furthermore, not every device will be online for every round of training realistically.

\subsubsection{Personalized Federated Learning} FedAvg generally fails as its objective is to find a single shared global model rather than specialized models for different groups of edge devices. Personalized FL methods, heavily based on Model Agnostic Meta Learning (MAML), run FedAvg followed by specialization \cite{Jiang2019ImprovingFL, Fallah2020PersonalizedFL}. Our approach eliminates many rounds of general model training by developing specialized models early on.

\section{Approach}
Algorithm 1 describes FedCD, which addresses non-IID federated learning with minimal communication and on-device memory overheads after convergence. FedCD clones high-performing models at milestone rounds and deletes low-performing models while updating model scores for each device.

The FedCD algorithm, like FedAvg, begins with a global model on a centralized server that all devices update to. At every milestone round, every model on the centralized server is cloned and compressed. 
In each training round, every participating device trains its local models for $E$ epochs, compresses the models, and sends its weight update and score (with some randomization) for each model to the global server, where each model's score on a given device reflects how well that model performs on the device's validation data. Then the server updates each global model by taking the weighted average of all devices' weight updates for that model and weighing them by that model's score. These global models are then re-deployed to the appropriate edge devices, and low-scoring models are deleted. 

\begin{algorithm}[h]
\caption{FedCD Algorithm.}
\label{alg}
\begin{algorithmic}
    \STATE {\bfseries Input:} Devices $i = 1, ..., N$, a global model $m=1$ ($M=1$)
    \STATE Initialize all scores $c_m^{(i)} = 1$
    \FOR{$t = 1, 2, \dots,T$}
            \STATE $\textit{round\_devices} =$ a random subset of $K$ devices
            \FOR {$i \in \textit{round\_devices}$}
                \STATE Device $i$ trains all models $m$ s.t. $c_m^{(i)} \neq 0$ on its local data for $E$ epochs
            \ENDFOR
            \FOR{$m = 1,2, \dots, M$}
                \STATE $\textit{w\_avg} = \text{AverageWeights}(i \text{ s.t. } c_m^{(i)} \neq 0)$\\
                \STATE Learner updates model $m$ with  \textit{w\_avg}\\
            \ENDFOR
            \STATE  Evaluate models with local validation data
            \STATE Update scores for all devices with normalized average of validation accuracy
            \STATE For each device $i$, delete underperforming models $m$ for which $\max(c^{(i)}) - c_m^{(i)} \geq \sigma(c^{(i)})$
            \STATE Delete models $m$ for which $c_m^{(i)} == 0$  for all devices $i$ from the central server
            \IF{$t$ is a milestone}
                \FOR{$m = 1, 2, \dots, M, i = 1, 2, \dots N$}
                    \IF{$c_m^{(i)} > 0$}
                        \STATE Clone model $m$ as model $M+m$ 
                    \ENDIF
                \ENDFOR
                \STATE $M = 2\cdot M$
                \STATE Normalize model scores for all devices
            \ENDIF
    \ENDFOR
\end{algorithmic}
\label{algo}
\end{algorithm}



Note that in Algorithm \ref{algo}, $M$ denotes the total number of previously created global models (including deleted models), which doubles at every milestone. Let $c_m^{(i)} \geq 0$ denote the score that device $i$ assigns model $m$, where a higher score denotes a better performing model.
We modify the weight update function as follows. Let $N$ be the number of devices. Let $w_m^{(i)}$ denote the weight vector for model $m$ by device $i$. Then we have
\begin{equation}
    w_m = {\sum_{i=1}^N w_m^{(i)} c_m^{(i)} \over \sum_{m =1}^M c_m^{(i)}}
\end{equation}

We experimentally investigated multiple ways of generating model score $c_m^{(i)}$ based on the accuracy $a_m^{(i)}[k]$ that model $m$ has on device $i$'s validation data in round $k$. 
We found that using a normalized average of the $\ell = 3$ most recent rounds' validation accuracy results in the highest performance while being robust to oscillation.  
Thus we define the score $c_m^{(i)}[r]$ of model $m$ by device $i$ at round $r$ as
\begin{align}
    s_m^{(i)}[r] &= \frac{\sum_{k=r-\ell}^{r-1} a_m^{(i)}[k]}{\ell}\\
    c_m^{(i)}[r] &= \frac{s_m^{(i)}[r]}{\sum_{m=1}^M s_m^{(i)}[r]}
\end{align}

When models are cloned, they receive the score of $1-c_p^{(i)}$, where $p$ denotes the parent model, to encourage differentiation between the parent models and the newly cloned models. 

To avoid exploding storage requirements, we delete all models $m$ for which the following holds
\begin{equation}
    \max(c^{(i)}) - c_m^{(i)} \geq \sigma(c^{(i)})
\end{equation}
where $\max(c^{(i)})$ denotes the score that device $i$ assigns to its highest performing model, and $\sigma(c^{(i)})$ denotes the standard deviation over the model scores by device $i$. Note that using a standard deviation based deletion criterion ensures that any device will maintain at least two models if there are at least two global models. After 20 rounds of training, if a device has two active models it will delete the lower-performing model $m'$ if $c_{m'}^{(i)} \leq 0.3$. 

For our experiments, we define the performance of a device as the accuracy of its highest-scoring model on its local testing data.

\subsection{Rationale}
By creating copies of the global model with different model scores to encourage exploration we can learn the archetypes of the edge devices and update weights based on the device's archetype. Then edge devices with the same archetype will preferentially update the same global model.

Each model fits its devices' distribution without access to the devices' data, thereby effectively addressing the problems that non-IID data pose to federated learning.
Compression via quantization allows for multiple smaller models on-device, and faster convergence leads to reduced communication cost.

\section{Experimental Results}
Our FedCD system consists of $30$ learners that have a non-IID subset of the global data.
To evaluate our approach, we compared the performance of FedCD to the performance of FedAvg on CIFAR-10, a dataset typically used for FL benchmarking, in two different setups. We also measured and compared the communication costs between the central server and devices and the time to convergence under FedCD and FedAvg.

\subsection{Setup}
We used data from CIFAR-10 \cite{CIFAR}, a comparison dataset standard for federated learning, consisting of 40k training images, 10k validation images, and 10k test images. Each device has a non-iid sample from the larger dataset that is consistent with its archetype to comprise its training/validation/test set. Each device received and sent weights to a 10-layer convolutional neural network. 
We exclusively used a device's validation set to determine its scores for a given model. We evaluated the best performing model for each device against its test set.\footnote{See https://github.com/jessijzhao/fedcd/ for code.}

Our experimental setup specified two required characteristics for each edge device: Archetypes (to describe the data distribution) and scores for each model (a normalized weighting of models that the device is maintaining). 15 devices participated in each training round and the global model was set to the weighted average of their updates.



\subsection{Hierarchical Archetypes}

In the real world, individual archetypes are seldom perfectly independent but rather can be grouped into "meta-archetypes" that each include several different archetypes.
An example of this structure are next-word predictions on phones of users living in a predominantly English-speaking country versus in a predominantly Spanish-speaking country (where the countries are meta-archetypes) of all ages (where the age groups are archetypes). Different age groups in the same country will likely share some common vernacular but common words across countries might be very limited due to the language barrier. 

To test the applicability of FedCD in this scenario, we constructed two sets of data (meta-archetypes that have data labeled 0,1,2,3,4 and 5,6,7,8,9 respectively) with 10 archetypes represented by the labels, i.e. an edge device of meta-archetype 1 only has access to training examples with labels 0, 1, 2, 3, and 4.
The experiment was run with 3 devices per archetype with bias $ b \sim Unif(0.6,0.7)$, where the bias denotes the fraction of a device's local dataset that consists of examples whose labels equal the archetype, i.e. a device with archetype 3 has $5k$ training images, of which $b* 5$k images have label $3$ and $(1-b)/4 * 5$k images have labels 0, 1, 2, and 4 each.
We set the cloning milestones at rounds 5, 15, 25, and 30. 

\begin{figure}[h]
\begin{subfigure}{\columnwidth}
\includegraphics[width=0.95\columnwidth, height=5cm]{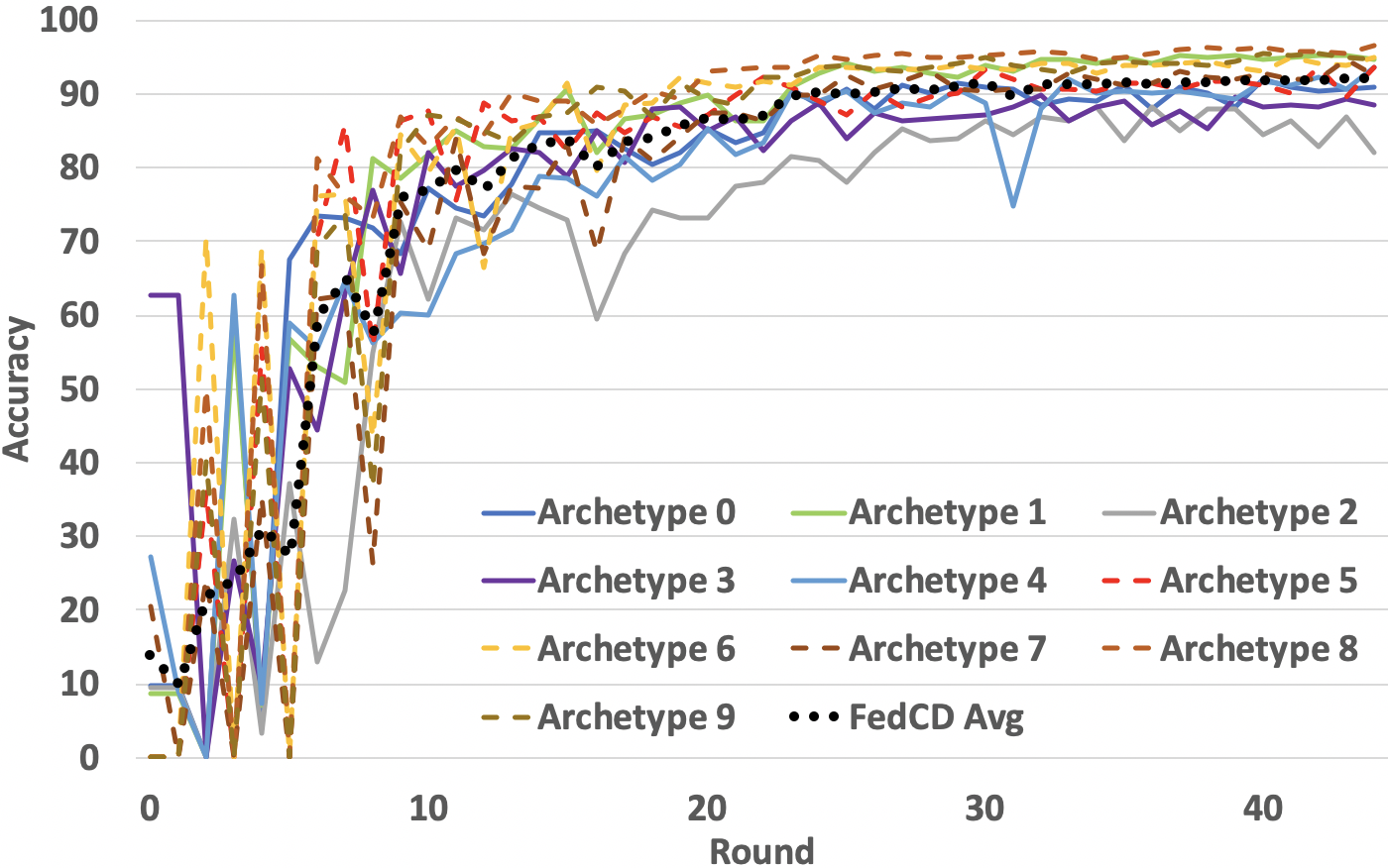} 
\caption{Test accuracy of the FedCD algorithm. There are 3 devices per archetype and their average is shown. Archetypes 0-4 belong to one meta-archetype and 5-9 to another.}
\label{fig:10_heirarch-FedCD}
\end{subfigure}%
\hfill
\begin{subfigure}{\columnwidth}
\includegraphics[width=0.95\columnwidth, height=6cm]{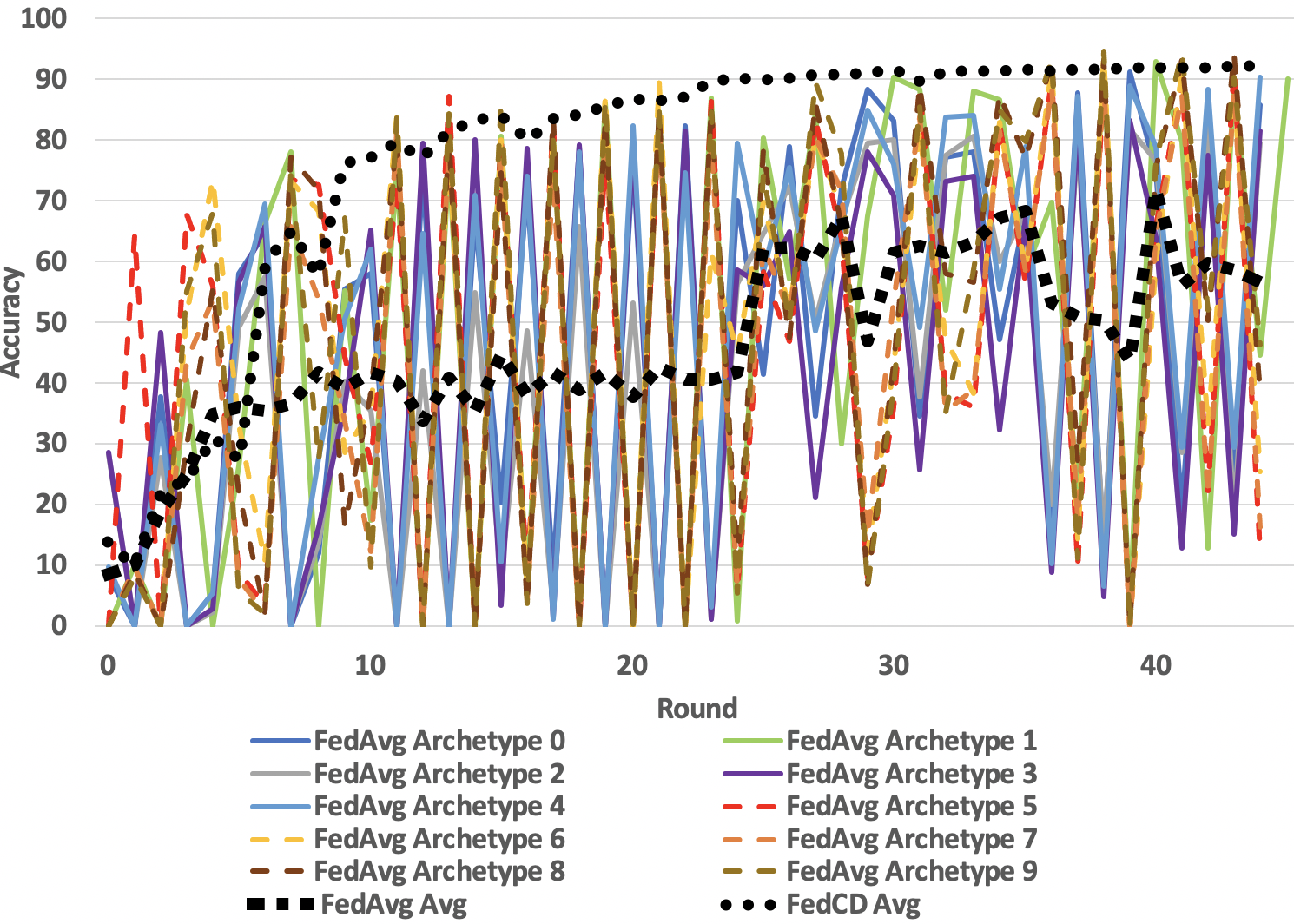}
\caption{Comparisons of test accuracy for the FedAvg and FedCD (dotted) algorithms over 50 rounds. FedAvg oscillates and underperforms FedCD.}
\label{fig:10_heirarch-baseline}
\end{subfigure}
\caption{Experiments with 10 archetypes within 2 meta-archetypes over 45 training rounds.}
\label{fig:10_heirarch}
\end{figure}

\begin{figure}[H]
    \centering
    \includegraphics[width=0.95\columnwidth]{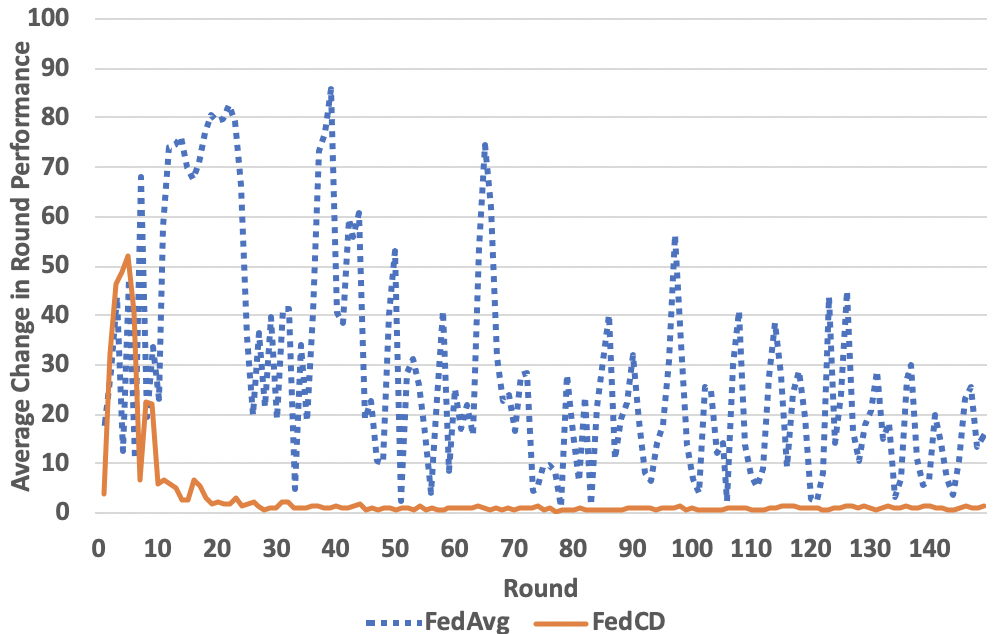}
    \caption{Average size of change in round-to-round performance of FedCD versus FedAvg across all devices for 150 rounds for the hierarchical archetypes experiment.}
    \label{fig:convergence-heirarch}
\end{figure}

Figure \ref{fig:10_heirarch} and \ref{fig:convergence-heirarch} show that FedCD converges relatively quickly (by round 35) and that FedCD is significantly more accurate on all archetypes than FedAvg. 

They also show that the two meta-archetypes converge to slightly different accuracies (meta-archetype 0, consisting of archetypes 0, 1, 2, 3, and 4, performs worse than meta-archetype 1, consisting of archetypes 5, 6, 7, 8, and 9.) 
The accuracy oscillations (where archetypes from the same meta-archetype oscillate together) in FedCD stop by round 10, whereas accuracy under FedAvg continues to oscillate past round 40 (see Figure \ref{fig:10_heirarch}). Furthermore, Figure \ref{fig:convergence-heirarch} shows that while FedCD converged after approximately 35 rounds, FedAvg failed to converge within 150 training rounds.

\subsection{Hypergeometric Archetypes}
Assuming a strict hierarchy excludes more complicated scenarios where the true distribution of data may be further or closer to two extremes. 
A real-world example of this setup are patient histories of citizens who visited hospitals across the US. In all parts of the country, an individual could have any disease, but hospitals in different locations may see different distributions of patients with respect to e.g. the severity of the disease, insurance quality, or socioeconomic status.

\begin{figure}[h]
    \centering
    \includegraphics[width=0.95\columnwidth]{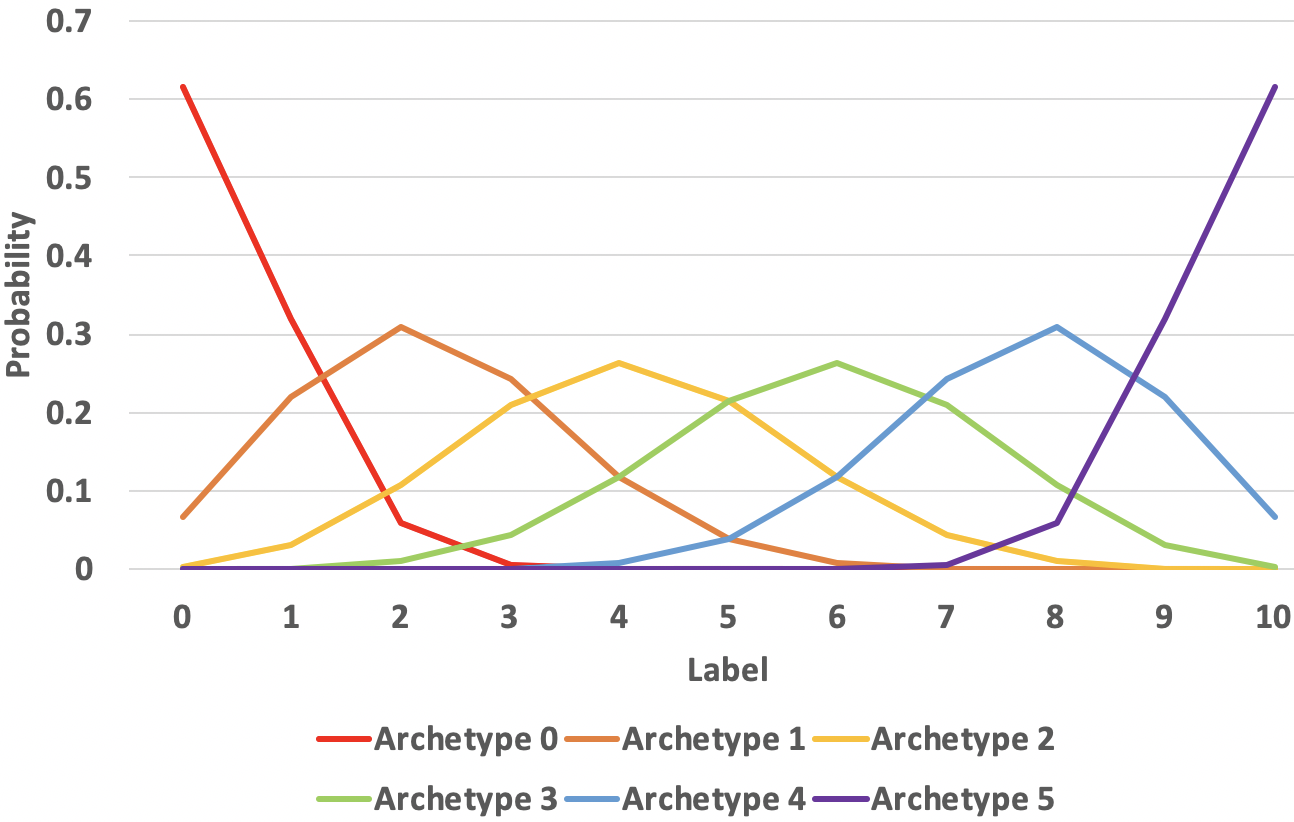}
    \caption{Visualization of the hypergeometric distribution for 6 archetypes across the 10 labels of CIFAR-10.}
    \label{fig:hypergeom_dist}
\end{figure}

To test the applicability of FedCD, each device sampled labeled training examples from a hypergeometric distribution over labels with $N=110, K\in \{5, 25, 45, 65, 85, 105\}$ based on its archetype, and $n = 10$ (see Figure \ref{fig:hypergeom_dist}).

We chose the hypergeometric distribution as it becomes a discrete approximation of the standard normal distribution when $N, K, n$ are large. Figure \ref{fig:hypergeom_dist} shows the data distribution for each archetype.
The experiment was run with 5 devices per archetype.

\begin{figure}[h]
\begin{subfigure}{\columnwidth}
\includegraphics[width=0.95\columnwidth, height=5cm]{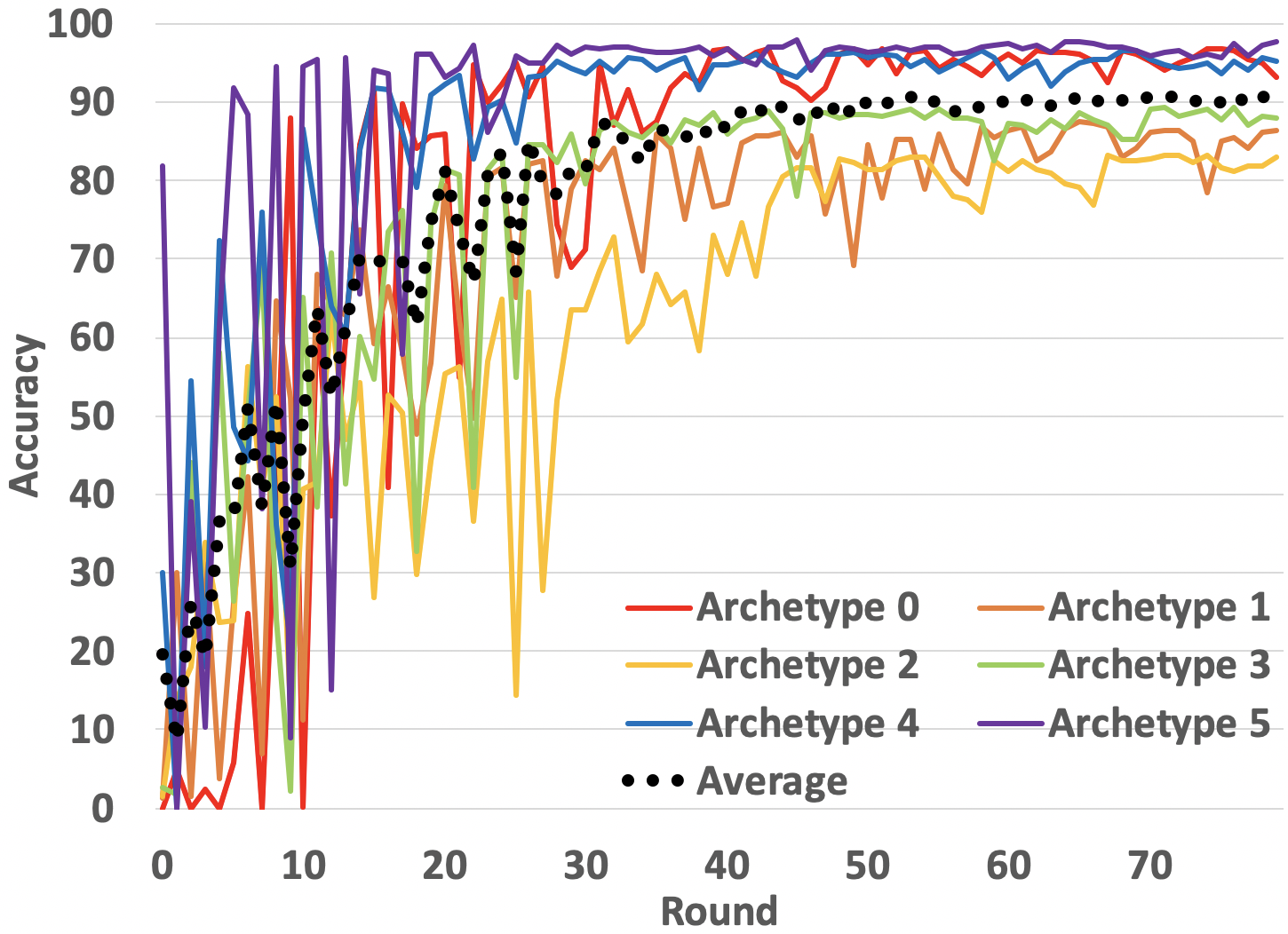} 
\caption{Test accuracy of the FedCD algorithm. There are 3 devices per archetype and their average is shown.}
\label{fig:6_hyper-FedCD}
\end{subfigure}%
\hfill
\begin{subfigure}{\columnwidth}
\includegraphics[width=0.95\columnwidth, height=5cm]{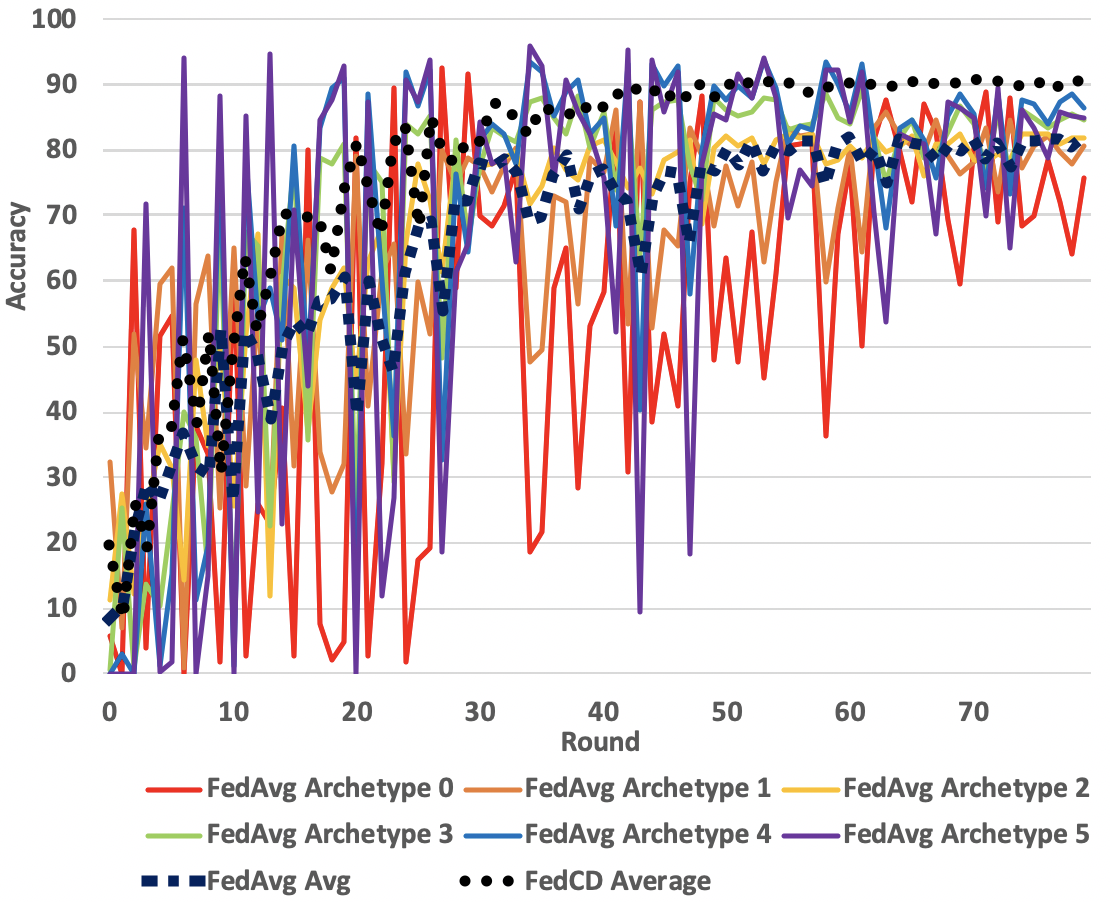}
\caption{Comparisons of test accuracy for the FedAvg and FedCD (dotted) algorithms.}
\label{fig:6_hyper-baseline}
\end{subfigure}
\caption{Experiments with 6 hypergeometric archetypes over 80 training rounds.}
\label{fig:6_hyper}
\end{figure}

We see in Figure \ref{fig:6_hyper-FedCD} that the FedCD algorithm converges quickly (by round 45) and that archetypes with more skewed probability distributions (archetypes whose distributions differ most from the global distribution, e.g. archetypes 0, 5) achieve higher accuracy than the central archetypes (archetypes whose distributions are most similar to the global distribution, e.g. archetypes 2, 3), since their distribution has a smaller standard deviation (see Figure \ref{fig:hypergeom_dist}).

\begin{figure}[H]
    \centering
    \includegraphics[width=0.95\columnwidth]{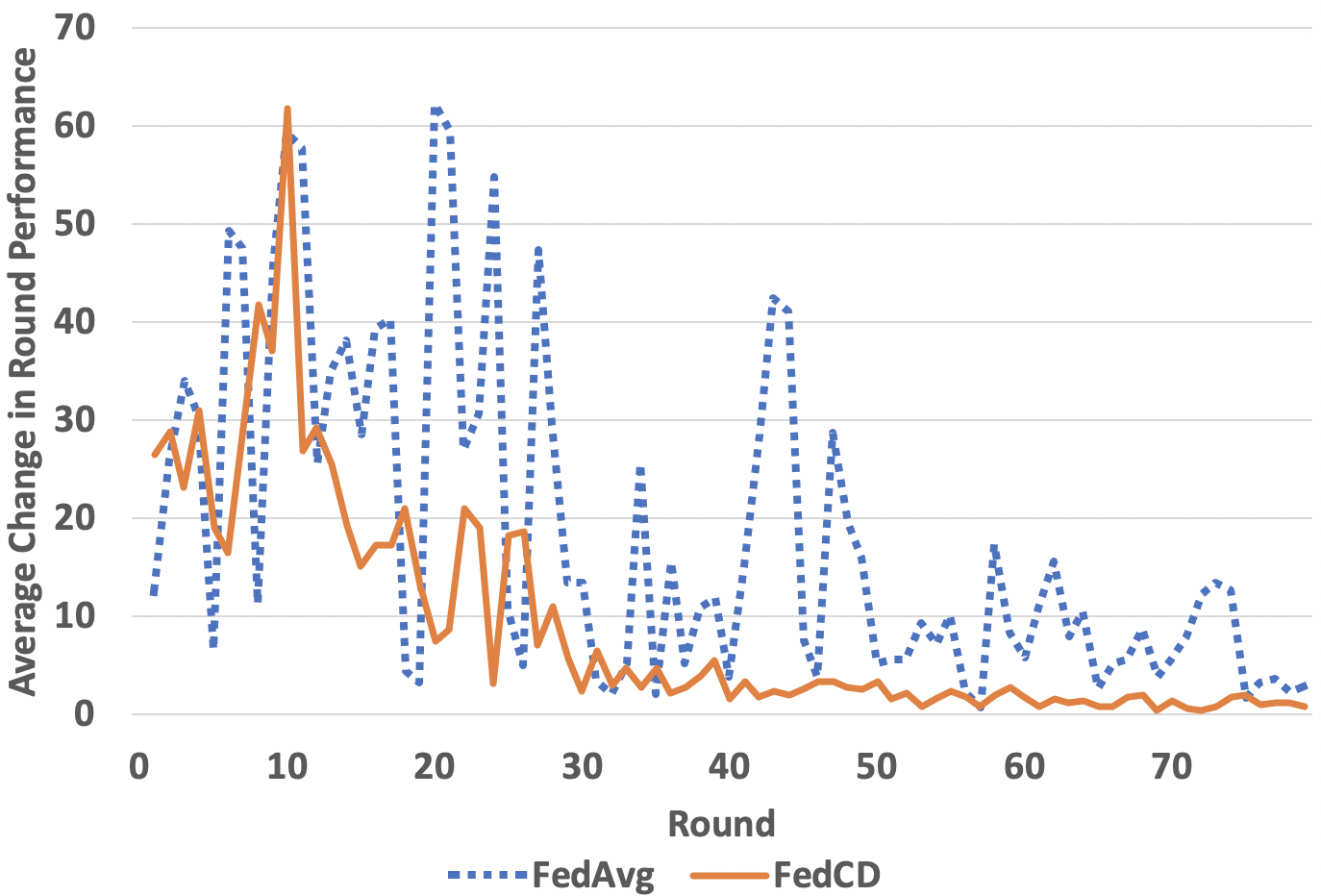}
    \caption{Average size of change in round-to-round performance of FedCD versus FedAvg on the hypergeometric archetypes experiment across all devices for 80 rounds.}
    \label{fig:convergence-hyper}
\end{figure}

Furthermore, while most archetypes converge under FedCD, many archetypes under FedAvg continue to oscillate as seen in Figure \ref{fig:6_hyper-baseline} as well as Figure \ref{fig:convergence-hyper}. 

In particular, while FedCD performs better on more skewed archetypes relative to other archetypes, FedAvg performs better and converges faster on the central archetypes. The increased success of FedCD on archetypes with more skewed data shows that FedCD indeed improves performance by learning specialized models that fit a given archetype's data distribution, as desired.

\subsection{Effects of Quantization}

\begin{figure}[H]
    \centering
    \includegraphics[width=0.95\columnwidth]{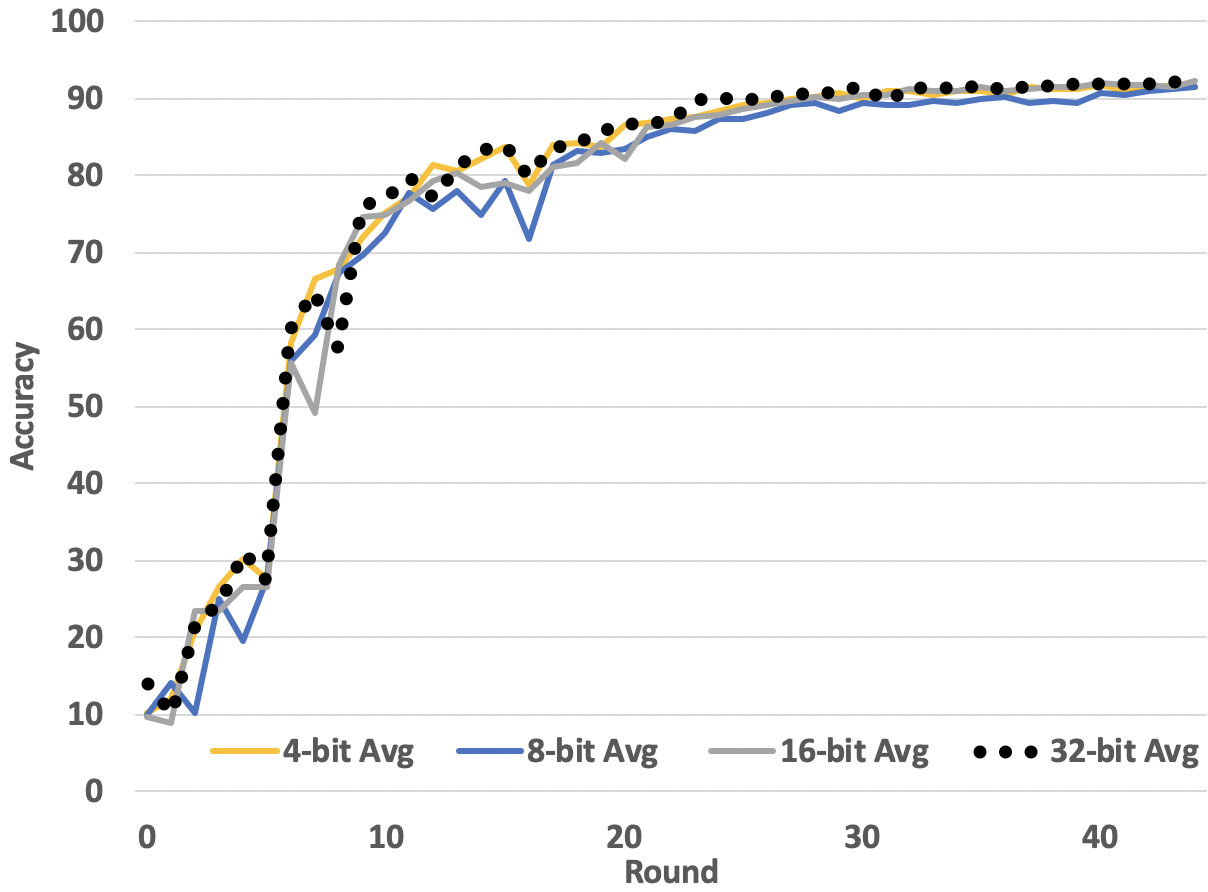}
    \caption{Effects of Quantization in the Hierarchical Experiment}
    \label{fig:quant1}
\end{figure}

Training multiple models on each device allows devices to self-sort into groups with similar archetypes by assigning similar scores to the same models. However, as on-device memory is limited, each model must be compressed to a smaller size, ideally without losing accuracy. 
 
Figure \ref{fig:quant1} shows that in the hierarchical archetypes experiment, different levels of quantization had no significant effect on model performance and only slightly impacted the time to convergence of the resulting models. Note that FedCD results in a single model per device, which is similarly insensitive to quantization as the FedAvg global model. Furthermore, while the central server may need to store significantly more models, relatively few models are maintained in practice.

\subsection{Model Selection Behavior}

Note that after $\ell$ rounds of cloning, there will exist at most $2^\ell$ global models. However, devices delete any models that already specialized for other archetypes as they will perform poorly on the device's data, such that these models are not cloned in future cloning rounds. Note that after 4 rounds of cloning, 10 out of 16 models were deleted from all devices.

\begin{figure}[h]
    \centering
    \includegraphics[width=0.8\columnwidth]{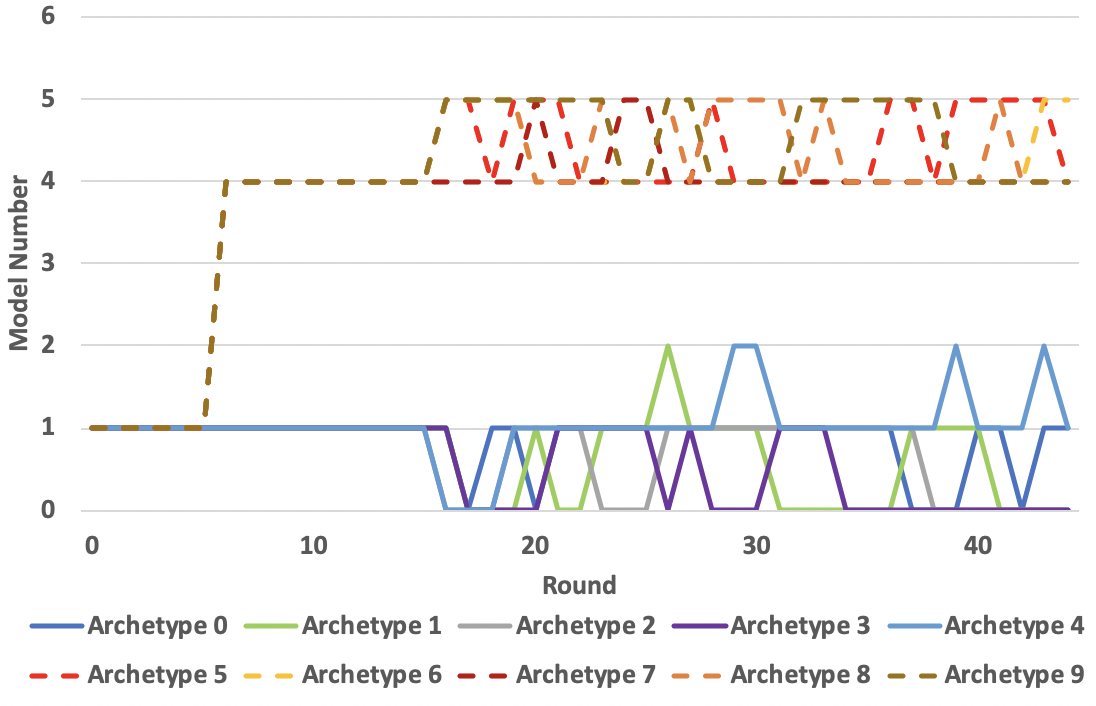}
    \caption{Model preference of different archetypes over the rounds of training for the hierarchical archetypes experiment in Figure \ref{fig:10_heirarch}.}
    \label{fig:correct}
\end{figure}

Figure \ref{fig:correct} depicts the consensus highest-scoring model that was not deleted by all devices for each archetype in the hierarchical archetypes experiment (consisting of 3 devices each). We can see that after the first cloning milestone at round 5, the devices segregate by meta-archetype. 
Subsequent cloning rounds have a limited effect, as the preferred model of individual archetypes oscillates between models 0 and 1 and models 4 and 5 respectively, indicating that these models perform similarly.


\subsection{Communication Costs}
Although the worst-case (each model is cloned at each milestone, i.e. $2^\ell$ models) would have an exponential communication cost overhead, devices tended to favor a single model and delete other models that didn't fit their data as well in practice. Note that this supposes the existence of archetypes (as in our experiments).

\begin{figure}
    \centering
    \includegraphics[width=0.95\columnwidth]{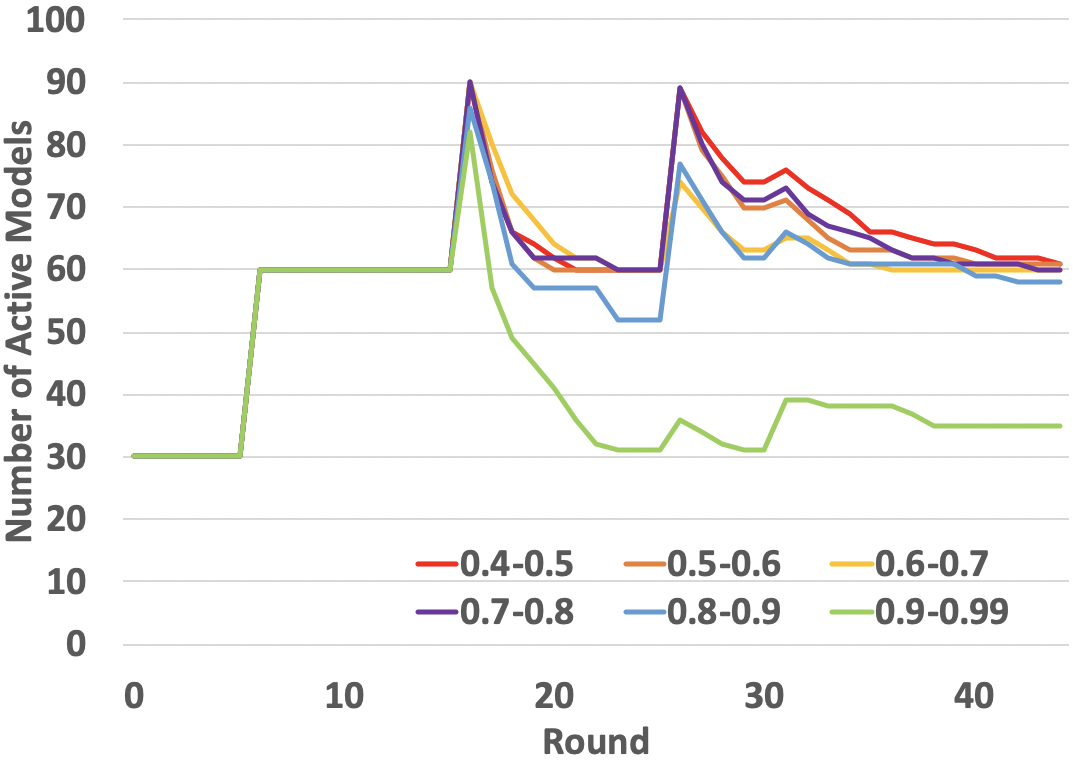}
    \caption{Total number of active models maintained across 30 devices over 45 training rounds of the hierarchical archetypes experiment with different device bias levels.}
    \label{fig:commcost}
\end{figure}

Figure \ref{fig:commcost} shows that the number of active models initially increases during the cloning rounds (5, 15, 25, 30) and drops during the subsequent rounds as devices delete models they no longer update to. In the end, each of the 30 devices update at most two active models and only a total of 6 models were preferred by any given device.

As the bias and therefore the difference between archetypes increases in the hierarchical archetypes experiment, devices of similar archetypes converge to similar models faster by scoring them higher than other models. In contrast, as the bias decreases and therefore the data of different archetypes becomes more similar (note that a bias of 0.2 represents the IID case within a meta-archetype), models become more similar as well such that devices tend to maintain multiple models for a larger number of rounds.

\begin{figure}[H]
    \centering
    \includegraphics[scale=0.45]{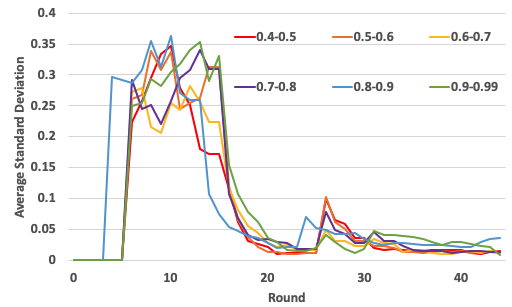}
    \caption{Average standard deviation of the 30 devices' scores (which sum to 1 for each device) over 45 rounds of the hierarchical archetypes experiment with different device bias levels. This shows that devices often end up with multiple models with similar scores.}
    \label{fig:std}
\end{figure}

The goal of FedCD is for each device to have one high-performing model and delete all other models. In some scenarios, such as the low-bias situations depicted in \ref{fig:commcost}, the algorithm terminates with each device having two equally-ranked high-performing models. This is fine as well, since each device can arbitrarily choose a model for deployment without loss of performance. Both cases would exhibit a low standard deviation of the scores they assign to active models (0 if all the scores were equal and 0 if there is a single model).
Figure \ref{fig:std} shows that the average standard deviation over model scores approaches 0 at the end of the training rounds for all levels of bias for the hierarchical archetype setup. 

\begin{table}
  \caption{Wall-clock comparison between FedAvg and FedCD}
  \label{tab:clock}
  \begin{tabular}{p{2.5cm}p{2.5cm}p{2.5cm}}
    \toprule
    Experiment& Rounds till \newline Convergence \newline (FedCD, FedAvg) &FedCD:FedAvg Wall-Clock Time\\
    \midrule
    Hypergeometric & 50,300*&1:3.488\\
    Hierarchical & 45,300* & 1:1.482\\
  \bottomrule
\end{tabular}
\end{table}

Table \ref{tab:clock} shows the wall-clock time for a run of FedCD versus a run of FedAvg till convergence. The run-time for FedAvg was capped at 300 rounds of training, since it had not converged by then for both the Hierarchical and Hypergeometric experiments. The wall-clock time of the experiments provide another insight into the advantage of FedCD as compared to the baseline, which takes a significant number of rounds to train.

\section{Conclusion} 

FedCD improves model performance on non-IID data by learning specialized models that best fit the data distribution of a group of similar devices (devices belonging to the same archetype).
Previous approaches have taken a decentralized approach by accepting complete peer-to-peer communication costs with full device participation in each round. However, this framework is sensitive to fluctuations in a real-world environment and incurs significant communication overhead.

Our centralized framework addresses these concerns by requiring only partial device participation in each round, though it incurs the costs of storing multiple quantized models on each device and the global server and sending multiple model updates per device during training.   
In this work, our main contributions are:
\begin{itemize}
    \item We propose a new framework for personalized FL.
    \item We empirically demonstrate that FedCD exhibits faster convergence and higher accuracy than the baseline FedAvg algorithm in several common non-iid scenarios. 
    \item We empirically show the number of active models (the total number of models stored on-device) does not explode by aggressively deleting poor-performing models from a local devices.
\end{itemize}


By amending the standard federated learning framework to train multiple global models simultaneously, we can improve model performance on non-IID data while incurring some limited communication and storage overhead during training.

\subsection{Future Work}

While we experimentally showed that FedCD converges faster and achieves higher accuracy at a reasonably low cost, future work could further analyze the dynamic nature of FedCD and attempt to find theoretical guarantees for convergence as well as bounds for communication and (server-side and on-device) storage costs.



Future work could also explore different types of bias other than label bias to determine the device archetypes, including archetypes defined by modifications to the input image. In addition, there are promising extensions of FedCD to other open problems in FL, such as using the cloning technique to address concerns regarding device bias and attack mitigation.

\section*{Acknowledgement}
We express our sincere appreciation to Professor H. T. Kung and Dr. Marcus Comiter for their valuable and constructive suggestions during the planning and development of this research. 
\bibliographystyle{ACM-Reference-Format}
\bibliography{references}
\end{document}